\documentclass[10pt,twocolumn,letterpaper]{article}

\usepackage{wacv}              

\usepackage{graphicx}
\usepackage{amsmath}
\usepackage{amssymb}
\usepackage{booktabs}
\usepackage{caption}
\usepackage{subcaption}
\usepackage{graphicx}
\usepackage{soul}
\usepackage{enumitem}
\usepackage{wrapfig}
\usepackage{tabularx}
\usepackage[dvipsnames]{xcolor}
\usepackage{multirow}
\newcommand{\method}{FedDiff}

\usepackage[pagebackref,breaklinks,colorlinks]{hyperref}

\usepackage[capitalize]{cleveref}
\crefname{section}{Sec.}{Secs.}
\Crefname{section}{Section}{Sections}
\Crefname{table}{Table}{Tables}
\crefname{table}{Tab.}{Tabs.}
\usepackage{url}

\def\wacvPaperID{799} 
\def\confName{WACV}
\def\confYear{2025}

\begin{document}

\title{Navigating Heterogeneity and Privacy in One-Shot Federated Learning \protect\\ with Diffusion Models}

\author{Mat\'ias Mendieta, Guangyu Sun, Chen Chen\\
Center for Research in Computer Vision, University of Central Florida, USA\\
{\tt\small \{matias.mendieta, guangyu\}@ucf.edu; chen.chen@crcv.ucf.edu}\\
}
\maketitle

\begin{abstract}
Federated learning (FL) enables multiple clients to train models collectively while preserving data privacy. However, FL faces challenges in terms of communication cost and data heterogeneity. One-shot federated learning has emerged as a solution by reducing communication rounds, improving efficiency, and providing better security against eavesdropping attacks. Nevertheless, data heterogeneity remains a significant challenge, impacting performance. This work explores the effectiveness of diffusion models in one-shot FL, demonstrating their applicability in addressing data heterogeneity and improving FL performance. Additionally, we investigate the utility of our diffusion model approach, FedDiff, compared to other one-shot FL methods under differential privacy (DP). Furthermore, to improve generated sample quality under DP settings, we propose a pragmatic Fourier Magnitude Filtering (FMF) method, enhancing the effectiveness of the generated data for global model training. Code available at \url{https://github.com/mmendiet/FedDiff}.
\end{abstract}

\section{Introduction}
Federated learning (FL) is a distributed machine learning technique that enables multiple clients to participate in the training process in a privacy-preserving manner. In FL, each client trains a local model on its own data and sends the model to a central server. The server combines these updates to improve the global model, which is then sent back to the clients. This approach ensures that the clients' data is kept private while enabling the central server to learn from the collective knowledge of all participating users \cite{openprob}.

However, FL poses significant challenges in terms of communication cost and data heterogeneity across clients. Communication cost is a major bottleneck in FL systems, as clients need to communicate frequently with the server over multiple rounds during the training process \cite{openprob, fedprox, feddyn}. This leads to a high communication overhead, making the process slow or infeasible. To overcome this challenge, \textbf{one-shot FL} has recently gained traction in the research community \cite{oneshot2019, dense, fl1, fl2}. In this setting, clients only communicate once with the server during the training process, significantly reducing the communication requirements. 
This approach not only improves the efficiency of the training process but also provides a better framework for privacy and application.
Specifically, one-shot FL provides better security against eavesdropping attacks, where adversaries attempt to steal or tamper with the information being sent between clients and the server \cite{eavesdropping}. By only requiring one round of communication, one-shot FL significantly reduces the likelihood of such attacks.
Furthermore, traditional multi-round training may not be a practical option in some cases, such as that of model markets \cite{fedkt}. In these scenarios, models are trained to be converged by a participating user and simply made available as a pretrained model to potential buyers without any option for iterative communication. 

However, another significant challenge in federated learning still remains, and that is the data heterogeneity problem \cite{fedalign, scaffold, moon, openprob}. 
In FL, clients often have very different data distributions, making optimization particularly challenging across the federated system.
In the one-shot setting, this is especially detrimental to performance. Without the luxury of multiple communication rounds, the resulting models will be significantly biased towards their narrow data distribution and difficult to reconcile into a global model. 
Knowledge distillation-based approaches have been studied in the literature in an attempt to address these problems \cite{oneshot2019, fedkt, dense}. 
Nonetheless, these methods still struggle immensely under high heterogeneity, resulting in large drops in performance.

Yet, another class of model is potentially well-suited for such heterogeneous distributions at the clients. Rather than simply employing discriminative models to train on the clients, one could instead leverage generative models. These generative models can then be gathered from the clients and inferenced on the server to form a dataset for global model training, eliminating the need for the challenging reconciliation process required for discriminative models. \cite{fedcvae} conducted a preliminary study of such a framework with conditional variational autoencoders (CVAEs) \cite{cvae} for one-shot FL, but there is still much to investigate in this paradigm. \textit{Specifically, we consider two primary research questions (RQ) in this work.}

\textit{\textbf{RQ1}}. We explore the utility of diffusion models in FL and their potential for improving the performance of the one-shot FL process. 
Diffusion models (DMs) \cite{ddpm} have emerged as prominent approaches for image generation, inspiring our investigation. In the trilemma between sample quality, diversity, and inference speed of generative models \cite{tri}, DMs excel in high sample quality and diversity. For one-shot FL, where inference occurs offline at the server, these two qualities where DMs excel are most crucial.
Therefore, we suggest that these and other specific traits of DMs are well-suited and could provide advantages for one-shot FL, as further detailed in Sec. \ref{sec:motive}.
We then validate this hypothesis through comprehensive experiments with our approach, \method, across various settings.

\textit{\textbf{RQ2}}. 
Generative models, while offering significant advantages for one-shot FL, can introduce privacy concerns. During generation, these models may produce images that are too similar to the original training data, potentially compromising data privacy.
Therefore, we investigate one-shot FL methods under provable privacy budgets with differential privacy (DP), 
as this aspect is not addressed by existing state-of-the-art one-shot FL works.
Safeguarding model privacy is critical in this setting since client models obtained in one-shot FL can be reused multiple times or traded in a model market.
Furthermore, in light of recent work \cite{diffmem}, we examine the potential memorization of diffusion models within our \method~approach and the effectiveness of DP as a mitigation strategy.
After studying these research questions, we further explore a simple technique for improving the performance of our \method~method under DP settings.
We observe that the quality of generated samples may deteriorate under DP constraints, rendering some samples counterproductive to the training of the global model. To improve the quality and consistency of the synthetic data, we propose a straightforward filtering approach, termed Fourier Magnitude Filtering (FMF). FMF leverages sample magnitudes derived from the Fourier transform to guide the selection of valuable samples. The resulting filtered dataset substantially improves the utility of the generated data, particularly in challenging conditions, as detailed in Sec. \ref{sec:fmf}.
Therefore, in this work, our \textbf{contributions} are summarized as follows:
\begin{itemize}[leftmargin=*,itemsep=2pt,topsep=0pt,parsep=0pt]
    \item We contribute to the FL literature with the first study exploring diffusion models in one-shot federated learning. 
    Our comprehensive investigation unveils the unique advantages inherent to diffusion models, enhancing the overall performance of one-shot FL while also addressing the significant challenges of data heterogeneity. We therefore establish a novel approach, \method, that not only ensures superior model performance but also aligns with the core requirements of one-shot FL.
    \item We study the privacy and utility of discriminative and generative-based SOTA one-shot FL methods with differential privacy (DP) guarantees under heterogeneous settings. Our \method~approach outperforms all other methods by a significant margin (from $\sim$5\% to $\sim$20\% across many datasets and settings), even when DP is employed.
    \item While \method~performs very well, we note that sample quality is affected under DP. Therefore, to improve performance in such conditions, we propose a simple Fourier Magnitude Filtering (FMF) approach, which improves the effectiveness of the generated data for global model training by removing low-quality samples. 
\end{itemize}

\section{Background and Preliminaries} \label{sec:background}

\textbf{One-shot Federated Learning.}
Federated learning (FL) has emerged as a promising paradigm for collaborative machine learning across decentralized devices while preserving data privacy. The seminal work by McMahan \etal~\cite{fedavg} introduced the concept of FL, where model updates are computed locally on user devices and aggregated on a central server. 
However, in the standard FL process, many iterative communication rounds are required for convergence. One-shot FL studies how to effectively learn in this distributed setting in a single round, thereby mitigating the need for many communication rounds. Several approaches have been proposed to tackle the unique characteristics of one-shot FL.
\cite{oneshot2019} introduce the one-shot federated learning framework and study several baseline approaches. In \cite{oneshot2019} and \cite{fedkt},  distillation approaches are studied using the ensemble of client models to train the global model, assuming a public dataset for this purpose. However, such an assumption is limited, as public data related to the domain of interest is often not available. A data-free method within the distillation methodology was proposed by \cite{dense}, where a generative adversarial network (GAN) is trained at the server level to generate the data for distillation, and iteratively optimized between distilling to the server model and training the GAN with the ensemble of client models. 

Nonetheless, these methods still struggle with heterogeneous environments, as we find in Sec. \ref{sec:motive}. Generative models on the client are well-suited for better undertaking in such settings, as they can focus on the narrow client distributions and simply generate data at the central location. \cite{fedcvae} introduce the use of CVAEs in highly heterogeneous one-shot FL. \ul{However, CVAEs exhibit suboptimal sample quality, a limitation that becomes markedly exacerbated with more complex datasets and when subjected to the constraints of DP, which are not explicitly addressed in the study by} \cite{fedcvae}.
In this work, we investigate diffusion models in one-shot FL and leverage their unique characteristics for the task, illustrating their potential in a variety of difficult FL settings and privacy guarantees.

\begin{figure*}[h]
\vspace{-4mm}
  \centering
  \includegraphics[trim={5 5 5 5},clip,width=\textwidth]{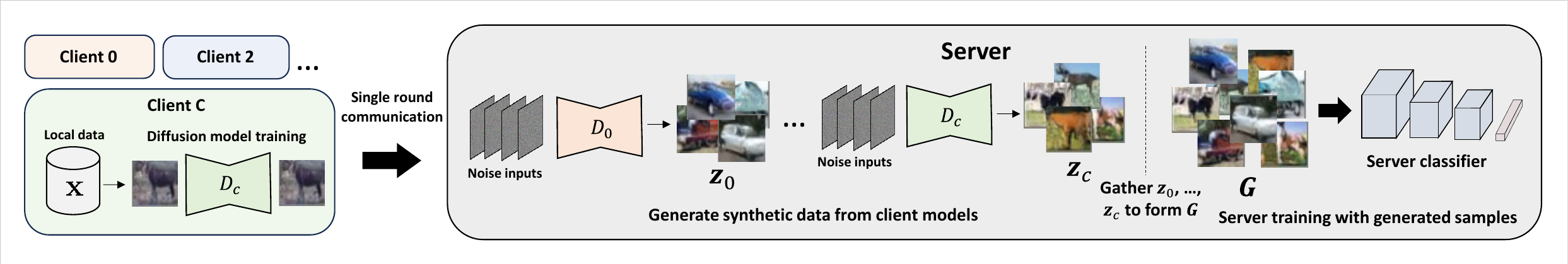}
  \vspace{-7mm}
  \caption{Our one-shot FL approach, \method. We first train a class-conditioned diffusion model on local data $\mathbf{x}$ at the clients. After completing training, the local diffusion models $D_{0}$, $D_{1}$, ..., $D_{c}$ are gathered by the server, where they are used to generate data $\mathbf{z}_0$, $\mathbf{z}_0$, ..., $\mathbf{z}_{c}$, which are combined to form the global training data $\mathbf{G}$. The global model is then trained on this synthetic dataset $\mathbf{G}$.}
  \vspace{-6mm}
  \label{fig:main}
\end{figure*}

\textbf{Diffusion Probabilistic Models.} Diffusion probabilistic models \cite{ddpm, beatgan}, or simply diffusion models as they are now commonly referenced (DM), have gained traction for application in generative vision tasks. Simply put, DMs aim to learn the backward process that can iteratively denoise an image corrupted with Gaussian noise back to the original. Specifically, as detailed in \cite{ddpm}, noise is introduced to a given sample via a Markovian chain forward process 
\vspace{-2mm}
\begin{equation}
    q(x_{1:T}|x_{0}) = \prod^T_{t=1}q(x_{t}|x_{t-1}),
    \vspace{-2mm}
\end{equation}
where $T$ is the total number of iterations (or timesteps) applied, and $q(x_{t}|x_{t-1})$ is parameterized by $\mathcal{N}(x_{t};\sqrt{1-\beta_{t}}x_{t-1}, \beta_{t}I)$. $\beta$ is a value between (0,1), and increases with timestep $t$, essentially making the final $q(x_{T}|x_{0})$ approximately a simple Gaussian $\mathcal{N}(0,I)$. This forward process is fixed, and the goal of the diffusion model is to learn the reverse process.
During training, we simply optimize for predicting the noise $\boldsymbol{\rho}$  from an arbitrary step $t$ in the forward process, forming a loss function \cite{ddpm}
\begin{equation}
L=\mathbb{E}_{t, \mathbf{x}_0, \boldsymbol{\rho}}\left[\left\|\boldsymbol{\rho}-\boldsymbol{\rho}_\theta\left(x_{t}, t\right)\right\|^2\right].
\end{equation}
 The process can also be conditioned on another variable $y$ in $\boldsymbol{\rho}_\theta\left(x_{t}, y, t\right)$. For example, the diffusion model can be class conditioned \cite{cfg}, with $y$ being a variable representing the class of the sample from a classification dataset. We utilize the class-conditioning approach of \cite{cfg} in our DMs for FL.

\textbf{Differential Privacy.} Differential privacy (DP) \cite{dp1,dp2,dp3} is a framework for ensuring that the output of a computation, such as machine learning model training, does not reveal sensitive information about any individual data point in the training dataset. A computation is said to be differentially private if the probability of obtaining a particular output is roughly the same whether a particular individual's data sample is included in the computation or not. Formally \cite{dp2}, 
\begin{equation}
Pr[A(D) \in S] \leq e^{\epsilon} \cdot Pr[A(D') \in S] + \delta,
\end{equation}
where $A$ is a randomized algorithm, $D$ and $D'$ are a pair of datasets that differ in at most one record, and $S$ is any subset of the output space of $A$. ($\epsilon, \delta$) control the level of privacy protection provided by the algorithm, essentially determining the maximum allowable amount of information that can be harnessed from the data. Larger values of $(\epsilon, \delta)$ correspond to weaker privacy guarantees, while smaller values of $(\epsilon, \delta)$ correspond to stronger guarantees. 

To train deep learning models with such guarantees, differentially private stochastic gradient descent (DPSGD) is typically employed \cite{dpsgd}.
In DPSGD, two main mechanisms are used to protect the privacy of individual data points: per-sample gradient clipping and the addition of random noise to the clipped gradients. Per-sample gradient clipping involves setting a maximum threshold on the norm of the gradient computed for each data point so that, if the norm of a gradient exceeds the threshold, it is rescaled. This step is necessary to limit the sensitivity of the loss function, which measures how much the loss function changes when a single data point is removed from the training dataset. After the gradients have been clipped, random noise is added to them before they are used to update the model parameters. The amount of noise added is calibrated based on privacy budget parameters $(\epsilon, \delta)$.

\section{Diffusion Models for Federated Learning} \label{sec:motive}
Before delving into the underlying motivation for our research questions \textit{\textbf{RQ1}} and \textit{\textbf{RQ2}}, it is essential to provide a brief exposition of the one-shot FL process when integrating generative models. The core premise of this approach departs from the traditional method of client-side discriminative model training. Instead, it advocates for the training of generative models on the client devices.
These client-side generative models are aggregated and used offline on the server side to synthesize data, which, in turn, facilitates the training of a global discriminative model. Within the scope of our study, we undertake an investigation into the viability of leveraging diffusion models in this paradigm.

\textbf{Why diffusion models?} In \cite{tri}, a generative learning trilemma is shown with model types, trading off sample quality, diversity, and fast sampling. CVAEs (as employed in \cite{fedcvae}) are typically identified to excel in diversity and fast sampling, but lacking in sample quality. However, for one-shot federated learning, fast sampling is not a concern, as the sampling can be done offline at the server (Figure \ref{fig:main}). Therefore, \textbf{\textit{high sample quality and diversity are more valuable properties in one-shot FL}}, as these will positively impact the performance of the trained global model with the synthetic data. In this trilemma, \textbf{\textit{diffusion models excel in sample quality and diversity}} \cite{tri}, but are not as quick to sample.
\ul{This motivated us to investigate the potential of DMs in this setting, as the inherent strengths of DMs align with the needs of one-shot FL.} 

Furthermore, while CVAEs and diffusion models share a common origin in terms of their objective, they differ in their approach to achieving this objective. 
The optimization task of the diffusion model is simplified to learning a Markov process to reverse a fixed forward process. The training is structured such that the model only needs to learn how to denoise a small step in the generation process, breaking down the problem.
In contrast, CVAEs must simultaneously learn both the forward process to encode the image to a latent space and the decoding process from that latent vector. We reason that the simplified objective of DMs helps achieve superior performance when dealing with complex data within the challenging FL environment (data heterogeneity, class imbalance, and limited sample sizes). Moreover, in the FL setting, privacy is of critical importance. To ensure privacy, training is done with DP, which introduces noise to the training process and increases the difficulty of optimization. In these settings, the simpler training paradigm of diffusion models is potentially advantageous. 

\textbf{Overview.} 
To provide a contextual foundation for our research inquiries, we start by laying out the settings of our study and approach in Sec. \ref{sec:setup}. With this groundwork, we investigate \textit{\textbf{RQ1}} in Sec. \ref{sec:rq1}, where we delve into the effectiveness of diffusion models in one-shot FL with our \method~approach. 
In Sec. \ref{sec:rq2}, we address \textit{\textbf{RQ2}} through a systematic exploration of one-shot FL methods within provable privacy budgets. We evaluate \method~and other SOTA approaches under DP constraints, as well as investigate the viability of DP in mitigating memorization. In Sec. \ref{sec:fmf}, we introduce our Fourier Magnitude Filtering approach, aimed at enhancing the efficacy of generated data for global model training by selectively eliminating low-quality samples.

\vspace{-2mm}
\subsection{\method~and Experimental Setup} \label{sec:setup}
\vspace{-2mm}
The basis of our \method~ is shown in Figure \ref{fig:main}. 
We begin by training class-conditioned diffusion models using the local data $\mathbf{x}$ on the clients. After training, the server collects these local models, denoted as $D_{0}$, $D_{1}$, ..., $D_{c}$, which are then used to generate data $\mathbf{z}_0$, $\mathbf{z}_{1}$, ..., $\mathbf{z}_{c}$. The label distributions from the clients are used to condition the generative models during generation, as in \cite{fedcvae}. The combination of these synthesized samples forms our global training dataset, $\mathbf{G}$. Subsequently, the global model is trained on the synthetic dataset $\mathbf{G}$ and evaluated in our experiments.

\textbf{Comparison Methods.} We compare with key baselines and the most recent state-of-the-art one-shot FL methods throughout our investigation. 

\textit{FedAvg} \cite{fedavg} is a standard baseline, which simply trains discriminative classifiers at the clients and averages their parameters, typically weighted by the number of samples at each client, to form a single server model.

\textit{DENSE} \cite{dense} is a one-shot FL approach that first trains the discriminative classifiers on the clients to convergence. Once the client models are collected, it performs two stages of training in an interactive manner, switching between training a GAN-based network for generating synthetic data and using the synthetic data to distill the ensemble of client models to a single server model. 

\textit{OneShot-Ens}. We also include an idealized variant of DENSE, where rather than attempt to distill the ensemble of client models to a single server model, we simply employ the ensemble as the final model, as shown in \cite{oneshot2019} and similarly compared to in \cite{fedcvae}. We term this approach OneShot-Ens throughout the paper.

\textit{FedCVAE} \cite{fedcvae}. This recently proposed method employs conditional variational autoencoders (CVAEs) for one-shot federated learning. Their approach has two variants, FedCVAE-KD and FedCVAE-Ens, which differ in how they operate at the server level. FedCVAE-KD distills all generative models from the clients to a single CVAE and then generates data for training the global model. On the other hand, FedCVAE-Ens employs each client model to generate data, contributing to the final dataset for training the server model. The latter variant always shows significantly better performance than the other in their paper; therefore, we compare with this FedCVAE-Ens variant and refer to it as FedCVAE in the rest of the paper.

\textbf{Datasets.} 
We employ three datasets, FashionMNIST \cite{fashion}, PathMNIST \cite{medmnistv2}, and CIFAR-10 \cite{cifar10}, which provide a range of domains and complexities. More details on the datasets are provided in the \textbf{Supp. Material}.
For our experiments, we divide the training set among $C$ clients with a Dirichlet distribution $Dir(\alpha)$, as commonly done in FL literature \cite{fedalign, dir1, dir2, fedcvae}. This partitioning approach creates imbalanced subsets, where some clients may not have any samples for certain classes. As a result, a significant number of clients will only encounter a small subset (or potentially only one) of the available class instances. We visualize data distributions with $Dir(0.1)$ and $Dir(0.001)$ across 10 clients in the \textbf{Supp. Material}.


\textbf{Federated Learning Settings.}  We reproduce DENSE and FedCVAE for our settings with their respective official code repositories. For all experiments, we perform 3 independent runs with different seeds and report the mean and standard deviation.
For all approaches, we train client models for 200 local epochs, as in \cite{dense}. For DENSE, FedCVAE, and \method, we train the final global model for 50 epochs.
For the generator of FedCVAE, we employ their CVAE variant with residual blocks, which has approximately 5.9M parameters. For our diffusion model, we employ a basic U-Net structure with residual blocks \cite{ddpm, unet} and class-conditioning, with similar parameters to FedCVAE ($\sim$5.8M). We employ a ResNet16 architecture for the discriminative models with approximately 6.4M parameters. For experiments with differential privacy, we employ the Opacus \cite{opacus} library in PyTorch \cite{pytorch} to track privacy budgets. Further details can be found in \textbf{Supp. Material}.

\begin{table}
\vspace{-2mm}
\caption{Data heterogeneity results with various $Dir(\alpha)$ partitions. Smaller $\alpha$ indicates higher levels of heterogeneity. Approaches leveraging discriminative models rapidly degrade in performance as heterogeneity increases. However, generative approaches are more robust to such conditions. \textbf{Our \method~shows superior performance to all, particularly in the most challenging scenarios (CIFAR-10, high heterogeneity)}.}

\label{tab:hetero}
\centering
\resizebox{0.46\textwidth}{!}{%
\begin{tabular}{ccccccc}
\toprule
& Method & $\alpha=0.1$ & $\alpha=0.01$ & $\alpha=0.001$ \\
\midrule
\multirow{5}{*}{FashionMNIST} & FedAvg & 57.11$\pm$3.64 & 29.50$\pm$10.6 & 25.89$\pm$4.78 \\
& DENSE & 65.20$\pm$3.55 & 28.92$\pm$17.3 & 27.68$\pm$4.08 \\
& OneShot-Ens & 67.35$\pm$1.19 & 33.79$\pm$17.9 & 32.01$\pm$3.35 \\
& FedCVAE & 78.08$\pm$2.69 & 78.81$\pm$3.25 & 81.53$\pm$0.23 \\
& \textbf{\method} & \textbf{87.21$\pm$0.74} & \textbf{86.81$\pm$0.54} & \textbf{86.59$\pm$0.69} \\
\midrule
\multirow{5}{*}{PathMNIST} & FedAvg & 28.10$\pm$4.60 & 22.05$\pm$8.20 & 21.92$\pm$4.95 \\
& DENSE & 50.97$\pm$3.19 & 29.26$\pm$10.7 & 27.69$\pm$4.52 \\
& OneShot-Ens & 34.62$\pm$3.61 & 34.94$\pm$9.32 & 34.49$\pm$5.30 \\
& FedCVAE & 41.60$\pm$0.82 & 44.81$\pm$1.41 & 47.35$\pm$3.21 \\
& \textbf{\method} & \textbf{74.58$\pm$1.02} & \textbf{70.61$\pm$1.37} & \textbf{69.43$\pm$1.30} \\
\midrule
\multirow{5}{*}{CIFAR-10} & FedAvg & 19.64$\pm$2.39 & 19.01$\pm$3.76 & 18.16$\pm$5.49 \\
& DENSE & 36.04$\pm$7.75 & 21.40$\pm$2.73 & 17.91$\pm$3.18 \\
& OneShot-Ens & 39.38$\pm$7.53 & 23.38$\pm$3.62 & 20.15$\pm$9.11 \\
& FedCVAE & 34.40$\pm$1.04 & 36.06$\pm$3.27 & 36.92$\pm$1.38 \\
& \textbf{\method} & \textbf{57.69$\pm$2.07} & \textbf{56.57$\pm$2.42} & \textbf{55.75$\pm$1.55} \\
\bottomrule
\end{tabular}%
}
\vspace{-5mm}
\end{table}
\begin{figure}[h]
     \centering
     \vspace{-1mm}
     \begin{subfigure}[]{0.22\textwidth}
         \centering
         \includegraphics[trim={0 0 0 0},clip,width=0.95\textwidth]{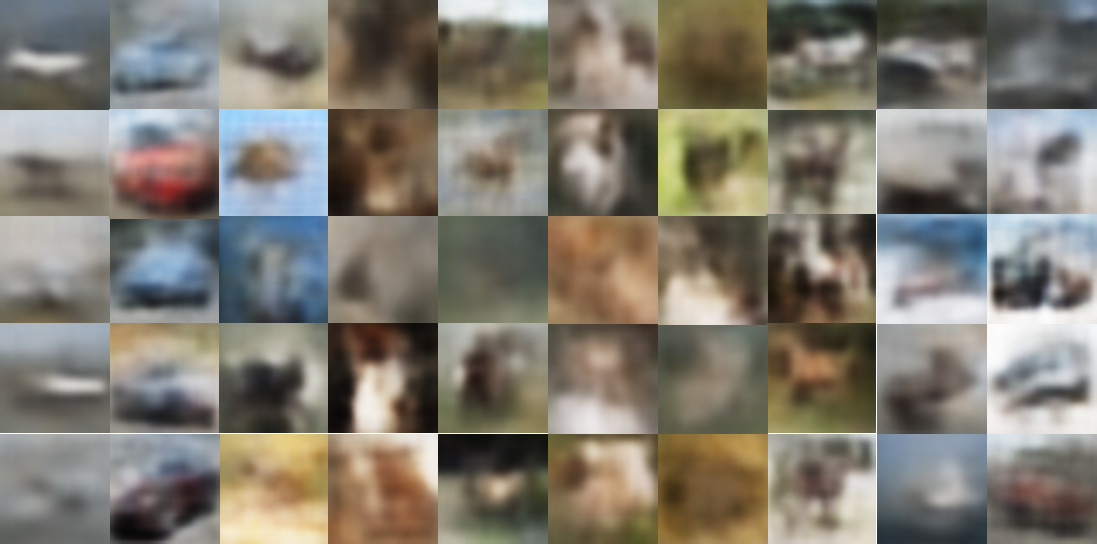}
         \caption{FedCVAE}
         \label{fig:y equals x}
     \end{subfigure}
     \begin{subfigure}[]{0.22\textwidth}
         \centering
         \includegraphics[trim={0 0 0 0},clip,width=0.95\textwidth]{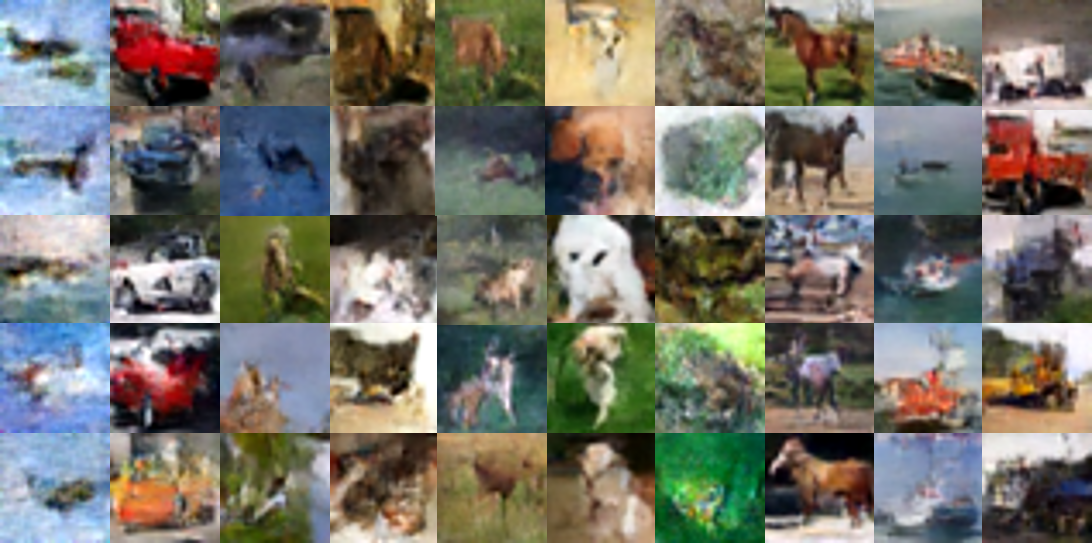}
         \caption{\method~(ours)}
         \label{fig:three sin x}
     \end{subfigure}
        \caption{Random sets of generated samples from FedCVAE and our \method. By leveraging the intrinsic properties of diffusion models, which are well-aligned with the requirements of one-shot FL, we achieve substantial benefits in sample quality and subsequent global model performance.}
        \label{fig:gen}
        \vspace{-5mm}
\end{figure}
\section{\textit{\textbf{RQ1}}: \method~for One-Shot FL} \label{sec:rq1}
We investigate \textit{\textbf{RQ1}} by exploring the efficacy of our \textit{\method}~ and other SOTA one-shot FL methods across important FL scenarios, including different data heterogeneity levels, number of clients, and resource requirements.
\subsection{Data Heterogeneity}
Data heterogeneity is a critical challenge in FL, particularly with one-shot settings. Even in the standard FL scenario of multiple communication rounds, client models often fit to very different distributions, and effectively reconciling their learnings is daunting. This is exacerbated in the one-shot setting, as we no longer have the luxury of getting many iterations to progressively steer the learning process towards an ideal encompassing representation.

In Table \ref{tab:hetero}, we analyze the performance of all methods under moderate ($Dir(0.1)$) to extreme ($Dir(0.001$) heterogeneity. Interestingly, \method~outperforms all other methods by a significant margin, from $\sim$5\% to up to $\sim$20\% in different scenarios.
In the case of CIFAR-10, which is the most complex of the datasets, we find that \method~provides the most improvement. As discussed in our initial motivations (Sec. \ref{sec:motive}), we reason that the focus on sample quality and diversity that is provided by the DM objective enables much-improved performance. Intuitively, this becomes increasingly evident in more complex settings. To verify this observation, we conduct a comparative analysis of generated samples produced by FedCVAE and our \method~approach, illustrated in Figure \ref{fig:gen}. The discernible disparity is evident, with the samples generated by our method exhibiting significantly enhanced sharpness and overall quality.

\begin{table}[]
\vspace{-2mm}
\caption{\hspace{-0.4mm}Results with varying number of clients $C$ with $Dir(0.01)$. As a fixed-size dataset is used in all experiments, increasing the number of clients also decreases the number of samples per client. We find that the SOTA discriminative approaches quickly degrade as the data is distributed across more clients. On the contrary, \textbf{our \method~maintains strong performance in all settings}.}
\label{tab:clients}
\centering
\resizebox{0.46\textwidth}{!}{%
\begin{tabular}{ccccccc}
\toprule
 & Method & $C=5$ & $C=10$ & $C=20$ \\
\midrule
\multirow{5}{*}{FashionMNIST} & FedAvg & 43.73$\pm$2.37 & 29.50$\pm$10.6 & 28.38$\pm$3.17 \\
& DENSE & 48.24$\pm$6.25 & 28.92$\pm$17.3 & 20.72$\pm$9.82 \\
& OneShot-Ens & 49.53$\pm$6.08 & 33.79$\pm$17.9 & 31.36$\pm$8.01 \\
& FedCVAE & 78.45$\pm$2.44 & 78.81$\pm$3.25 & 78.33$\pm$2.45 \\
& \textbf{\method} & \textbf{86.89$\pm$0.34} & \textbf{86.81$\pm$0.54} & \textbf{87.24$\pm$0.57} \\
\midrule
\multirow{5}{*}{PathAMNIST} & FedAvg & 29.18$\pm$3.54 & 22.05$\pm$8.20 & 19.83$\pm$3.16 \\
& DENSE & 33.39$\pm$6.14 & 29.26$\pm$10.7 & 20.79$\pm$5.77 \\
& OneShot-Ens & 36.95$\pm$5.30 & 34.94$\pm$9.32 & 24.59$\pm$6.19 \\
& FedCVAE & 46.16$\pm$1.17 & 44.81$\pm$1.41 & 41.25$\pm$1.68 \\
& \textbf{\method} & \textbf{72.74$\pm$0.63} & \textbf{70.61$\pm$1.37} & \textbf{69.11$\pm$0.99} \\
\midrule
\multirow{5}{*}{CIFAR-10} & FedAvg & 29.14$\pm$5.15 & 19.24$\pm$3.77 & 15.79$\pm$2.64 \\
& DENSE & 30.48$\pm$2.30 & 21.40$\pm$2.73 & 12.60$\pm$2.33 \\
& OneShot-Ens & 36.17$\pm$3.21 & 23.38$\pm$3.62 & 13.23$\pm$2.96 \\
& FedCVAE & 32.34$\pm$2.59 & 36.06$\pm$3.27 & 37.63$\pm$1.87 \\
& \textbf{\method} & \textbf{57.68$\pm$1.86} & \textbf{56.57$\pm$2.42} & \textbf{58.45$\pm$0.73} \\
\bottomrule
\end{tabular}%
}
\vspace{-5mm}
\end{table}
\vspace{-1mm}
\subsection{Number of Clients}
\vspace{-1mm}
We also study the effect of the number of clients $C$ in Table \ref{tab:clients}. As we employ the same total number of samples in all experiments, the number of samples \textit{per client} will increase with smaller $C$, and decrease with larger $C$. This allows us to observe the effect of increasing the distributed nature of the data across the client network.

One question arising from the adoption of generative models in FL settings pertains to their ability to maintain satisfactory performance when trained on a limited number of samples.
Interestingly, when analyzing the results, we find that \method~is capable of handling a much smaller number of client training samples with little to no performance degradation. On the other hand, the discriminative model approaches quickly experience a collapse in performance when expanding to 20 clients. In the heterogeneous environment of federated learning, the local optimization of a discriminative model on a highly imbalanced and small dataset proves challenging. Rather than being an overwhelming burden, such a situation is handled well by \method, as its sole focus is to capture the subsequently smaller distribution. Furthermore, we again find that \method~outperforms FedCVAE in all settings, further illustrating the potential for diffusion models in one-shot FL.

\subsection{Resource Requirements}
\vspace{-1mm}
To further explore the efficacy of our method, we also examine resource factors, including FLOPs and parameter count, for each method deployed on a single client.
Notably, our \method~approach consistently delivers superior accuracy with comparable computational resources to other methods. We extend this assessment to a reduced model size (\method$_{s}$ in Table \ref{tab:resource}), reaffirming its strong performance relative to alternative methods. This analysis underscores the effectiveness of \method, even when deployed on hardware with modest computational capabilities.

\begin{table} 
\vspace{-2mm}
\small
\caption{Accuracy versus FLOPs and parameter count (Params) for each method on a single client. Our \method~approach consistently attains heightened accuracy levels while maintaining very reasonable resource demands on par with other methodologies. We also evaluate our method with a scaled-down model variant (\method$_{s}$), further confirming its performance relative to alternative approaches. This analysis underscores the realistic feasibility of our \method~framework.}
\label{tab:resource}
\centering
\resizebox{0.47\textwidth}{!}{%
\begin{tabular}{ccc@{\hspace{7pt}}|@{\hspace{5pt}}ccc}
\toprule
\multicolumn{1}{c}{}&\multicolumn{2}{c}{Resources} & \multicolumn{3}{c}{Accuracy} \\
\cmidrule(r){2-3} \cmidrule(l){4-6}
 Method & MFLOPs $\downarrow$ & Params $\downarrow$ & FashionMNIST & PathMNIST & CIFAR-10 \\
\toprule
FedAvg &  479.92 & 6.44M & 29.50$\pm$10.6 & 22.05$\pm$8.20 & 19.24$\pm$3.77 \\
DENSE &  479.92 & 6.44M & 28.92$\pm$17.3 & 29.26$\pm$10.7 & 21.40$\pm$2.73 \\
OneShot-Ens & 479.92 & 6.44M & 33.79$\pm$17.9 & 34.94$\pm$9.32 & 23.38$\pm$3.62 \\
FedCVAE &  79.00 & 5.97M & 78.81$\pm$3.25 & 44.81$\pm$1.41 & 36.06$\pm$3.27 \\
\textbf{\method} &  301.14 & 5.81M & \textbf{86.81$\pm$0.54} & \textbf{70.61$\pm$1.37} & \textbf{56.57$\pm$2.42} \\
\textbf{\method$_{s}$} &  \textbf{77.43} & \textbf{1.46M} & \textbf{85.90$\pm$0.92} & \textbf{70.53$\pm$5.61} & \textbf{50.08$\pm$1.87} \\
\bottomrule
\end{tabular}
}
\vspace{-7mm}
\end{table}
It is pertinent to emphasize that training diffusion models within the \method~framework is \textit{no more intricate than conventional methodologies and remains highly viable for FL}. The computational complexity aligns with training a conventional CNN model with a modest number of parameters, and we employ the same number of local epochs as previous work with CNNs \cite{dense}. Importantly, the training process entails selecting random steps in the diffusion process at any given training iteration, eliminating the necessity for sequential steps during training. During inference, the generation process involves sequential denoising steps; however, this poses no issue for the clients, as generation occurs at the server in \method. 
Therefore, \method~is an effective and practical approach for providing strong performance. Further discussions on communication and server-side operations are provided in the \textbf{Supplementary Material}.
\vspace{-1mm}

\section{\textit{\textbf{RQ2}}: Privacy Considerations} \label{sec:rq2}
\vspace{-1mm}

Privacy holds paramount importance in one-shot FL. The trained client model may be repeatedly utilized or even exchanged in a model market context, and therefore, safeguarding the privacy of the model before it leaves the client is imperative. 
\ul{However, other SOTA works have not experimented with DP constraints, nor have they thoroughly explored this aspect, often leaving privacy discussions simply as a possibility for future work} \cite{dense, fedcvae}. In the subsequent sections, we meticulously investigate privacy from various perspectives and dive into our research question \textit{\textbf{RQ2}}. Furthermore, we propose a novel approach in Sec. \ref{sec:fmf} to improve performance under strict privacy constraints.
\vspace{-3mm}

\subsection{Differential Privacy}\label{sec:exdp}
\vspace{-2mm}
Differential privacy is the widely accepted standard for ensuring privacy of a model, as it offers a provable guarantee of privacy \cite{dpsgd, dp1, dp2, dp3, dpgold}. Utilizing $(\epsilon, \delta)$ DP during model training guarantees comprehensive privacy protection, encompassing not only the model's parameters and activations but also extending to all subsequent downstream operations. \ul{It is important to note that a model trained under $(\epsilon, \delta)$ DP safeguards the privacy of every training sample, regardless of its qualities or uniqueness} \cite{opacus}. 
Specifically, we train all approaches under ($\epsilon, \delta$) DP at the clients for various privacy levels of $\epsilon$ = 50, 25, and 10, with $\delta=10e^{-5}$, $C=10$, and $\alpha=0.01$. Lower $\epsilon$ values correspond to a tighter privacy budget, and the stated budget is for the entire training of each local model. We employ the Opacus \cite{opacus} library for implementing DP. 
Further DP training details are provided in \textbf{Supp. Material}. We present the results in Table \ref{tab:dp}. 

\begin{table}
\vspace{-2mm}
\caption{Differential privacy (DP) results under various $\epsilon$ budgets. We set $C=10$ and $\alpha=0.01$ as the default setting. \textbf{Even under DP constraints, \method~is a particularly viable approach, outperforming all other SOTA one-shot FL methods}.}
\label{tab:dp}
\centering
\resizebox{0.86\columnwidth}{!}{%
\begin{tabular}{ccccccc}
\toprule
& Method & $\epsilon=50$ & $\epsilon=25$ & $\epsilon=10$ \\
\midrule
\multirow{5}{*}{FashionMNIST} & FedAvg & 21.04$\pm$12.1 & 20.82$\pm$12.3 & 20.39$\pm$12.6 \\
& DENSE & 26.34$\pm$9.03 & 26.29$\pm$9.81 & 24.29$\pm$15.6 \\
& OneShot-Ens & 31.27$\pm$10.9 & 31.32$\pm$10.1 & 29.99$\pm$16.7 \\
& FedCVAE & 44.40$\pm$1.70 & 43.89$\pm$2.53 & 41.65$\pm$3.19 \\
& \textbf{\method} & \textbf{75.92$\pm$1.86} & \textbf{75.08$\pm$2.13} & \textbf{73.43$\pm$1.50} \\
\midrule
\multirow{5}{*}{PathMNIST} & FedAvg & 16.98$\pm$8.93 & 15.30$\pm$6.44 & 14.85$\pm$4.19 \\
& DENSE & 20.56$\pm$6.59 & 19.19$\pm$3.76 & 18.41$\pm$1.86 \\
& OneShot-Ens & 24.59$\pm$7.63 & 23.38$\pm$2.60 & 22.23$\pm$2.02 \\
& FedCVAE & 24.06$\pm$1.57 & 22.15$\pm$2.68 & 20.51$\pm$1.29 \\
& \textbf{\method} & \textbf{54.98$\pm$2.04} & \textbf{51.51$\pm$1.85} & \textbf{47.85$\pm$3.68} \\
\midrule
\multirow{5}{*}{CIFAR-10} & FedAvg & 16.35$\pm$1.52 & 15.39$\pm$1.87 & 15.07$\pm$2.12 \\
& DENSE & 16.97$\pm$2.35 & 15.68$\pm$2.27 & 14.98$\pm$1.25 \\
& OneShot-Ens & 17.73$\pm$2.71 & 17.34$\pm$2.35 & 15.72$\pm$1.34 \\
& FedCVAE & 16.29$\pm$1.55 & 16.08$\pm$2.19 & 15.86$\pm$2.83 \\
& \textbf{\method} & \textbf{32.93$\pm$1.93} & \textbf{31.76$\pm$2.68} & \textbf{27.78$\pm$1.66} \\
\bottomrule
\end{tabular}
}
\vspace{-6mm}
\end{table}
As expected, all methods experience a drop in performance when trained under DP settings. Nonetheless, \method~still stands out, outperforming all other methods by a significant margin. Particularly for FashionMNIST, \method~experiences comparatively less accuracy drop under DP than FedCVAE. 
As articulated in our initial motivations outlined in Section \ref{sec:motive}, 
DP training introduces noise into the training process, exacerbating the complexity of optimization. In such scenarios, the simplicity of the training paradigm employed by diffusion models becomes notably advantageous. 
To demonstrate this even further, we conduct an experiment with an even tighter privacy budget of $\epsilon = 1$ for FedDiff, achieving $65.53\pm0.70$, $44.38\pm3.35$, and $21.48\pm1.53$ on FashionMNIST, PathMNIST, and CIFAR-10 respectively. As other methods already have very low accuracy at $\epsilon=10$, restricting any further would be counterproductive. On the other hand, despite facing even more stringent privacy constraints, \textit{\method~maintains a higher level of performance at $\epsilon$ = 1 than all other methods in Table \ref{tab:dp} at $\epsilon$ = 50}.
Overall, we show that \method~is a strong approach even when DP is employed.
\begin{figure*}[h]
\vspace{-7mm}
     \centering
     \begin{subfigure}[]{0.33\textwidth}
         \centering
         \includegraphics[width=0.93\textwidth]{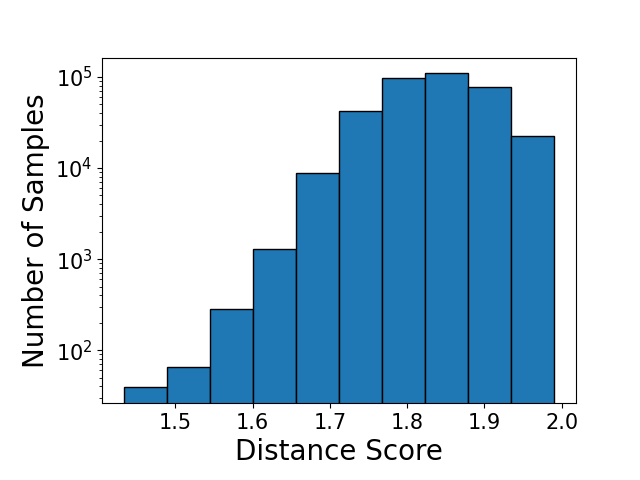}
         \caption{FashionMNIST}
     \end{subfigure}
     \hfill
      \begin{subfigure}[]{0.33\textwidth}
         \centering
         \includegraphics[width=0.93\textwidth]{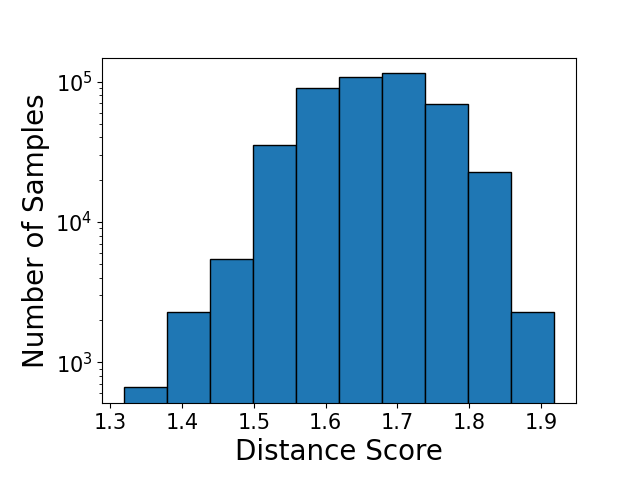}
         \caption{PathMNIST}
     \end{subfigure}
     \hfill
     \begin{subfigure}[]{0.33\textwidth}
         \centering
         \includegraphics[width=0.93\textwidth]{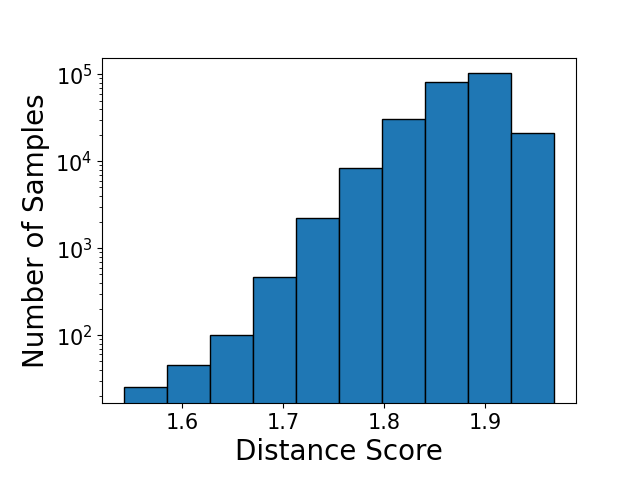}
         \caption{CIFAR-10}
     \end{subfigure}
        \caption{Histogram of distance scores for all generated samples at $\epsilon=50$ to corresponding closest training image by Eq. \ref{eq:mem} on each dataset. Note that the y-axis is in \textit{log scale}, as there are very few samples with lower scores.}
        \label{fig:hist}
\end{figure*}
\begin{figure*}[h]
\vspace{-2mm}
     \centering
     \begin{subfigure}[]{0.28\textwidth}
         \centering
         \includegraphics[width=0.92\textwidth]{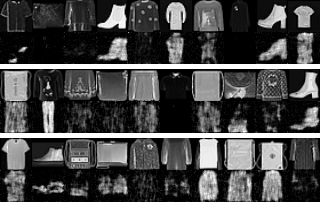}
         \caption{FashionMNIST}
     \end{subfigure}
     \hspace{5mm}
      \begin{subfigure}[]{0.28\textwidth}
         \centering
         \includegraphics[width=0.92\textwidth]{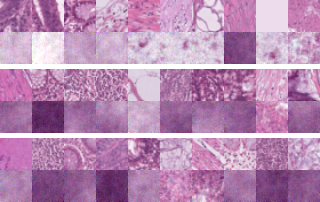}
         \caption{PathMNIST}
     \end{subfigure}
     \hspace{5mm}
     \begin{subfigure}[]{0.28\textwidth}
         \centering
         \includegraphics[width=0.92\textwidth]{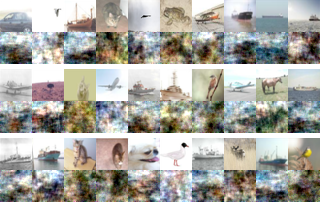}
         \caption{CIFAR-10}
     \end{subfigure}
        \caption{Qualitative comparison of original training samples and generated samples at $\epsilon=50$. We show the closest 30 samples via the similarity metric in Equation \ref{eq:mem}. In each stacked row, the original samples are on top, with the corresponding nearest generated image immediately below. Even under the loosest privacy guarantee of $\epsilon=50$, we do not see blatant memorization.}
        \label{fig:mem}
        \vspace{-5.5mm}
\end{figure*}
\subsection{Addressing Memorization} \label{sec:mem}
\vspace{-2mm}
In federated learning, one of the most common attacks to violate the privacy of client models and reconstruct their local training data is to use gradient inversion \cite{seethrough, gradinv2}, where gradient information from the trained client model is exploited. However, existing literature lacks studies on performing gradient inversion on diffusion models, designing their attacks for discriminative architectures. Nonetheless, this omission is well-founded, as a more logical approach is to generate data directly via diffusion model sampling. Therefore, a crucial question arises: do DMs trained in a one-shot FL setting memorize training samples and regenerate them during inference? A recent study by \cite{diffmem} explored diffusion models in centralized settings and identified their potential to memorize samples under specific conditions.
They acknowledged differential privacy as the gold standard defense strategy, but did not provide completed experiments. 
\ul{Therefore, we evaluate the effectiveness of DP to this end, assessing memorization within our DP-trained models to investigate whether inadvertent reproduction of the training data can be eliminated.}
To conduct this study, we adopt the evaluation methodology established by \cite{diffmem} to scrutinize the occurrence of memorization. Specifically, from each DP-trained diffusion model, we generate a vast number of samples (five times the size of the training set). Subsequently, for each generated image, we assess potential memorization compared to the original training samples using the adaptive distance metric introduced by \cite{diffmem},
\vspace{-2mm}
\begin{equation} \label{eq:mem}
\ell\left(\hat{x}, x ; S_{\hat{x}}\right)=\frac{\ell_2(\hat{x}, x)}{\alpha \cdot \mathbb{E}_{y \in S{\hat{x}}}\left[\ell_2(\hat{x}, y)\right]}.
\vspace{-2mm}
\end{equation}
Here, $S_{\hat{x}}$ denotes the set comprising the $n$ nearest elements from the training dataset to the example $\hat{x}$. The resulting distance metric yields a small value if the extracted image $x$ exhibits significantly closer proximity to the training image $\hat{x}$ compared to the $n$ closest neighbors of $\hat{x}$ within the training set. The idea is to find generated images that are unusually close to an original training image as an indication of memorization. We set $\alpha=0.5$ and $n=50$ as in \cite{diffmem}. \cite{diffmem} did not define the specific threshold for Equation \ref{eq:mem} for marking when a sample is considered memorized. Therefore, we consider the intuitive threshold to be less than 1, as this would indicate that the distance from the extracted image to the training image is less than half of the average distance to the closest $n$ neighbors. Upon conducting this assessment, we do not find any instances of memorized samples for all datasets under such definition, even at an elevated privacy parameter of $\epsilon=50$, with the closest distance values being $\sim$1.3. 
We show the histogram of scores for all samples on each dataset in Figure \ref{fig:hist}. Because the threshold definition for memorization could vary, we also qualitatively show the samples with the lowest distances for all datasets at $\epsilon=50$ in Figure \ref{fig:mem}. Notably, the training versus the generated samples have discernible differences, in contrast to the nearly identical samples uncovered in \cite{diffmem} when training large DMs without DP. Also, given the nature of FL, the choice of diffusion model size will typically be small (for example, ours is $\sim$5.8M parameters), and therefore will be less likely to memorize compared to the larger DMs evaluated in \cite{diffmem}.
\vspace{-2mm}
\subsection{Fourier Magnitude Filtering} \label{sec:fmf}
\begin{figure*}[t]
     \vspace{-4mm}
     \centering
     \begin{subfigure}[]{0.33\textwidth}
         \centering
         \includegraphics[trim={0 0 0 20},clip,width=0.88\textwidth]{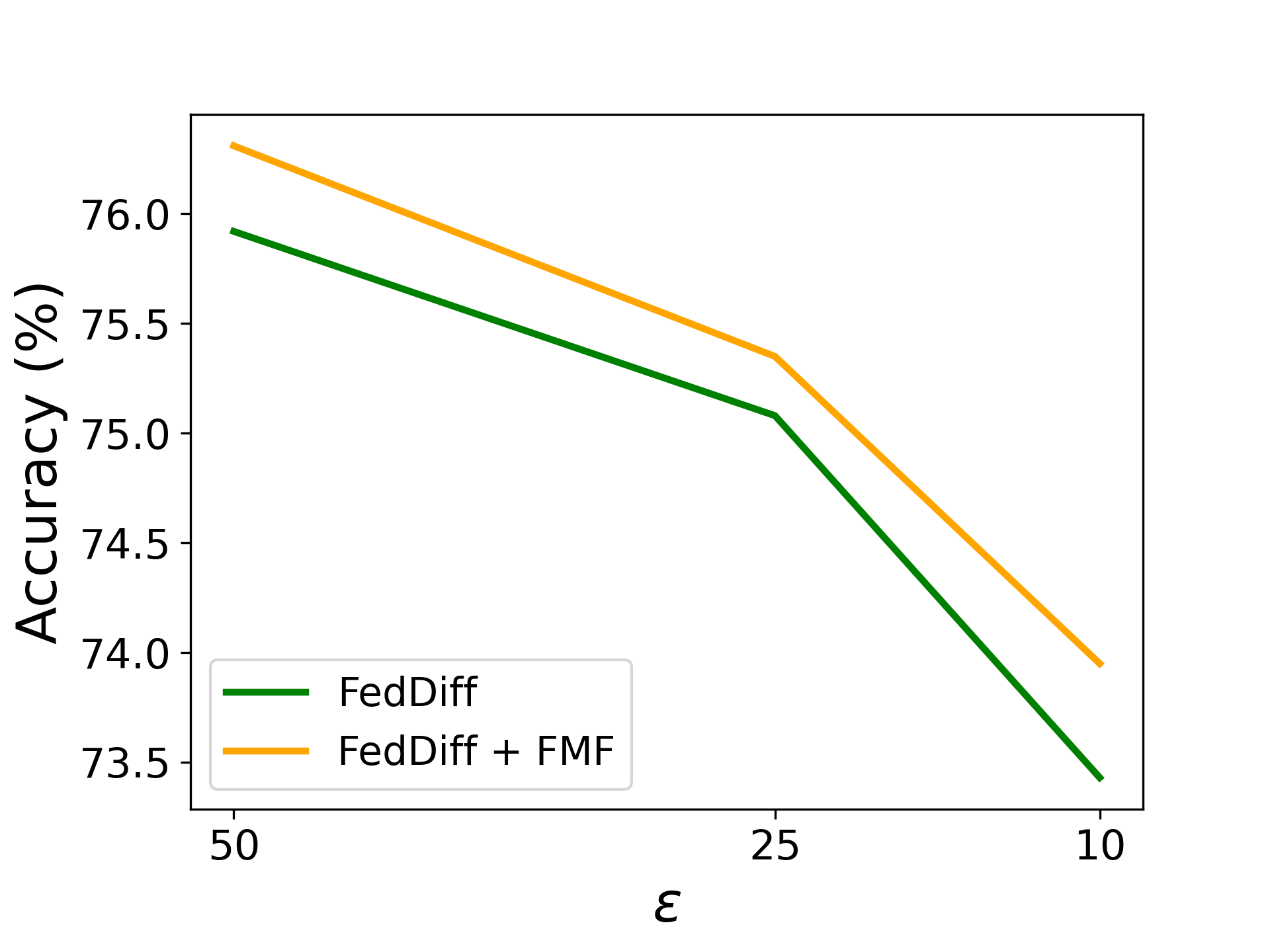}
         \caption{FashionMNIST}
     \end{subfigure}
     \hfill
      \begin{subfigure}[]{0.33\textwidth}
         \centering
         \includegraphics[trim={0 0 0 20},clip,width=0.88\textwidth]{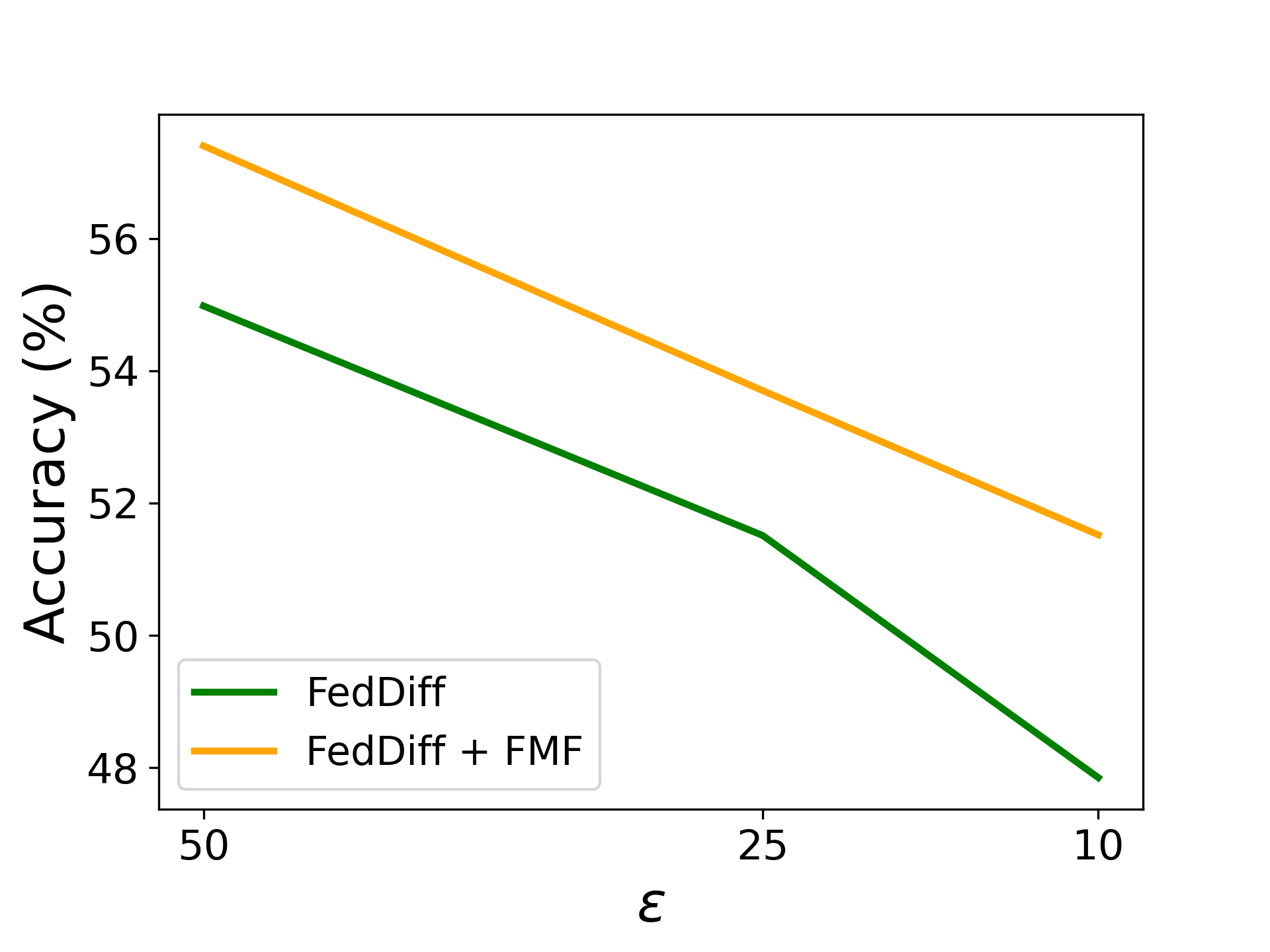}
         \caption{PathMNSIT}
     \end{subfigure}
     \hfill
     \begin{subfigure}[]{0.33\textwidth}
         \centering
         \includegraphics[trim={0 0 0 20},clip,width=0.88\textwidth]{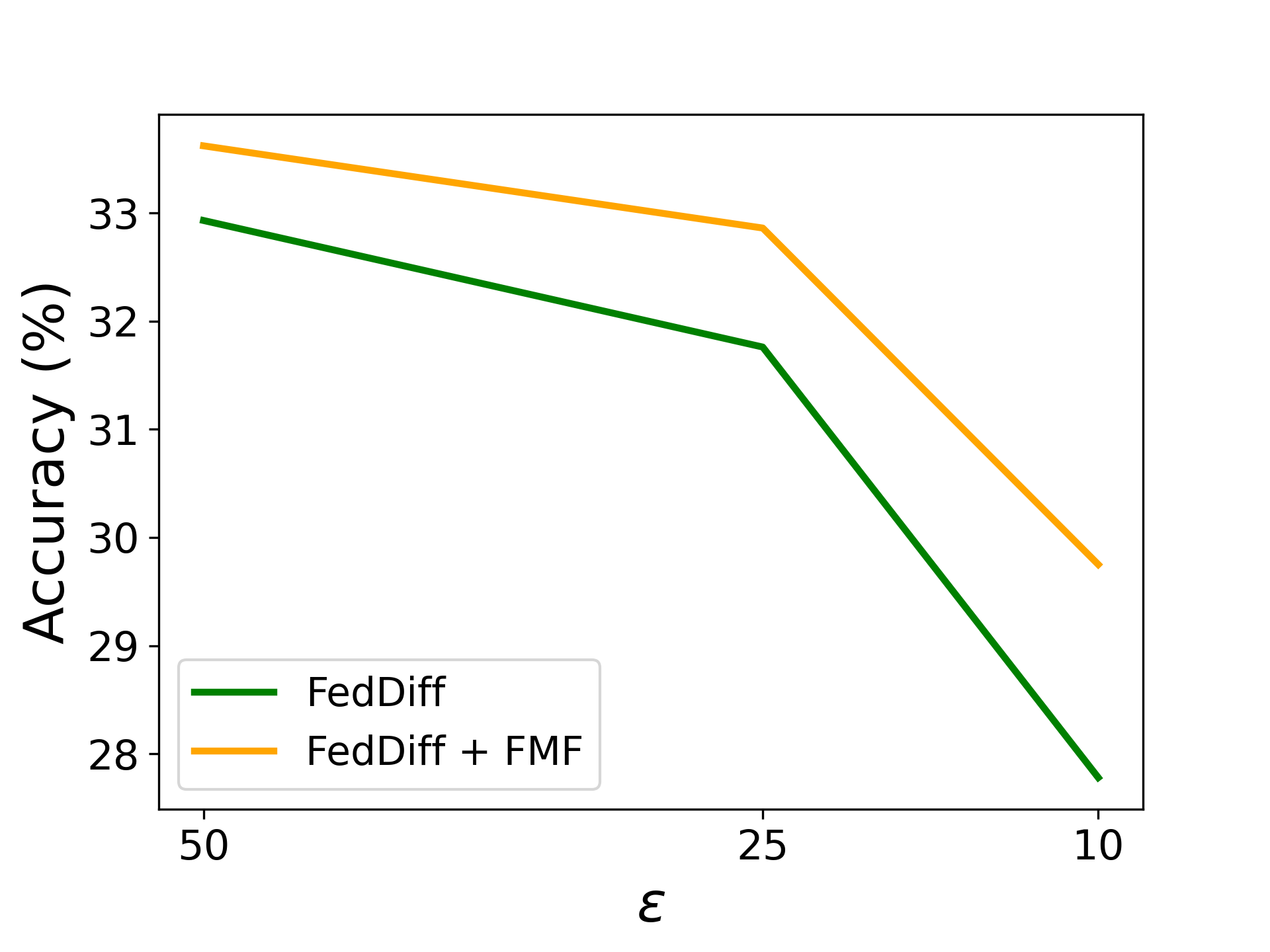}
         \caption{CIFAR-10}
     \end{subfigure}
        \caption{Results with our Fourier Magnitude Filtering under DP. \method~is in \textbf{\textcolor{ForestGreen}{green}} and \method+FMF in \textbf{\textcolor{orange}{orange}}. Our FMF approach provides a simple way to boost accuracy, especially in more challenging scenarios such as lower $\epsilon$ budgets and more complex datasets. We plot the mean across three runs with different seeds for each setting. Additional $\gamma$ ablations are provided in the \textbf{Supplementary Material}.}
        \label{fig:fmf}
        \vspace{-5mm}
\end{figure*}
\vspace{-2mm}
While \method~performs comparatively well against other SOTA one-shot FL methods under DP constraints, we further investigate a simple approach to improve our method,
particularly for complex data most affected by DP. 
As shown in Figure \ref{fig:mem}, we note that the generated samples under DP can lack details, exhibiting reduced structure.
Therefore, it may be advantageous to sort out and remove such poor quality samples from the final synthetic dataset prior to conducting the training of the global model.


To understand the impact of prioritizing data quality on performance, we conduct an initial experiment. For CIFAR-10, we leverage a centralized pretrained classifier as an oracle to discern high and low quality samples. We selectively retain samples for which the oracle accurately classifies and discard those it misclassifies. This provides a way to filter out samples that are likely irrelevant or misleading for training a model. We then train the global model exclusively on the curated dataset of accurate samples and evaluate. This investigation yields a discernible improvement ranging from approximately 2\% to 4\% in final global model accuracy compared to training with all generated data, verifying the importance of data quality. Therefore, a critical question arises from this observation: \textit{how can we conduct sample filtration in the absence of an oracle?}

To do so, we look to the Fourier domain for a potential source of information. We note that the magnitude of local client images has been utilized in FL to assist in domain generalization across clients by providing low-level ``style" information without the high-level semantics encoded in the phase \cite{feddg}. In our case, we propose to leverage the Fourier magnitude information as a potential referenceable indicator to guide the sample filtering process. 

On the client, we take the Fourier transform of the local samples and extract the magnitude information. For each client $c$, we gather the average sample magnitude with 
\vspace{-2mm}
\begin{equation}\label{eq:mag}
\textstyle
    \bar{M_{c}} = \frac{1}{n_{c}} \sum_{i=1}^{n_{c}} |\psi(\mathbf{x}^{i})|,
\vspace{-2mm}
\end{equation}
where $\psi$ is the 2D Fourier transform operation, $\mathbf{x}^{i}$ is a sample, and $n_{c}$ is the total number of samples in client $c$.
$\bar{M}$ is bundled with the model and transmitted by the client to the relevant global party.
As in our standard global training procedure, samples are generated with the client-trained diffusion models to form a synthetic set. Prior to conducting global training, we calculate a sample score $s$ for the generated data $z$ from each diffusion model from the clients, $s_{\mathbf{z}^{i}_{c}} = \||\psi(\mathbf{z}^{i}_{c})| - \bar{M}_{c}\|_2$.
We can then leverage this information to guide the removal of irrelevant samples, forming the final training set $\mathbf{G}$ by removing $\gamma$ percent of the generated data with the highest $s$ (larger magnitude difference).
To continually ensure privacy guarantees, we apply DP in the FMF calculation. We do so by employing the DP bounded mean \cite{dpmean} from PyDP~\cite{pydp} to calculate the average magnitude $\bar{M_{c}}$ at each client. This allows us to precisely manage any degree of privacy leakage for $\bar{M_{c}}$ and include it in the overall privacy budget.

In Figure \ref{fig:fmf}, we show the results of applying our FMF approach with \method~for the same overall DP budgets as Table \ref{tab:dp}. FMF is particularly effective in the most difficult scenarios, helping to mitigate the performance drop in harsh FL environments. For example, FMF provides over 3.5\% and 2\% improvements with PathMNIST and CIFAR-10 in the challenging $\epsilon=10$ setting. Overall, FMF is a simple way to boost performance in one-shot FL under DP.
\vspace{-2mm}

\section{Conclusion} \label{sec:conclusion}
\vspace{-1mm}
Our work addresses two valuable research questions in one-shot FL. Firstly, we investigate the potential of DMs for one-shot FL and present the pioneering effort in this direction. 
We unveil the unique advantages that DMs offer, showcasing their potential to enhance performance and tackle heterogeneity across diverse settings with our proposed approach, FedDiff.
Secondly, we study privacy in SOTA one-shot FL and contribute a thorough investigation under provable privacy budgets, as well as address memorization concerns.
Furthermore, to enhance performance under harsh DP conditions, we propose a novel and pragmatic solution, Fourier Magnitude Filtering, to boost the efficacy of generated data for global model training by eliminating low-quality samples. More discussions, including limitations and broader impact, and additional experiments are included in the \textbf{Supp. Material}. We hope our work will inspire the community and foster further research in this direction to improve one-shot FL with generative models.

\noindent\textbf{Acknowledgment.} This work is partially supported by the NSF/Intel Partnership on MLWiNS under Grant No. 2003198.

{\small
\bibliographystyle{ieee_fullname}
\bibliography{egbib}
}

\clearpage
\appendix
\section*{Supplementary Material} 
\section{Additional Training Details} \label{sec:train}
\begin{figure}[h]
     \begin{subfigure}[]{0.49\columnwidth}
         \centering
         \includegraphics[width=\columnwidth]{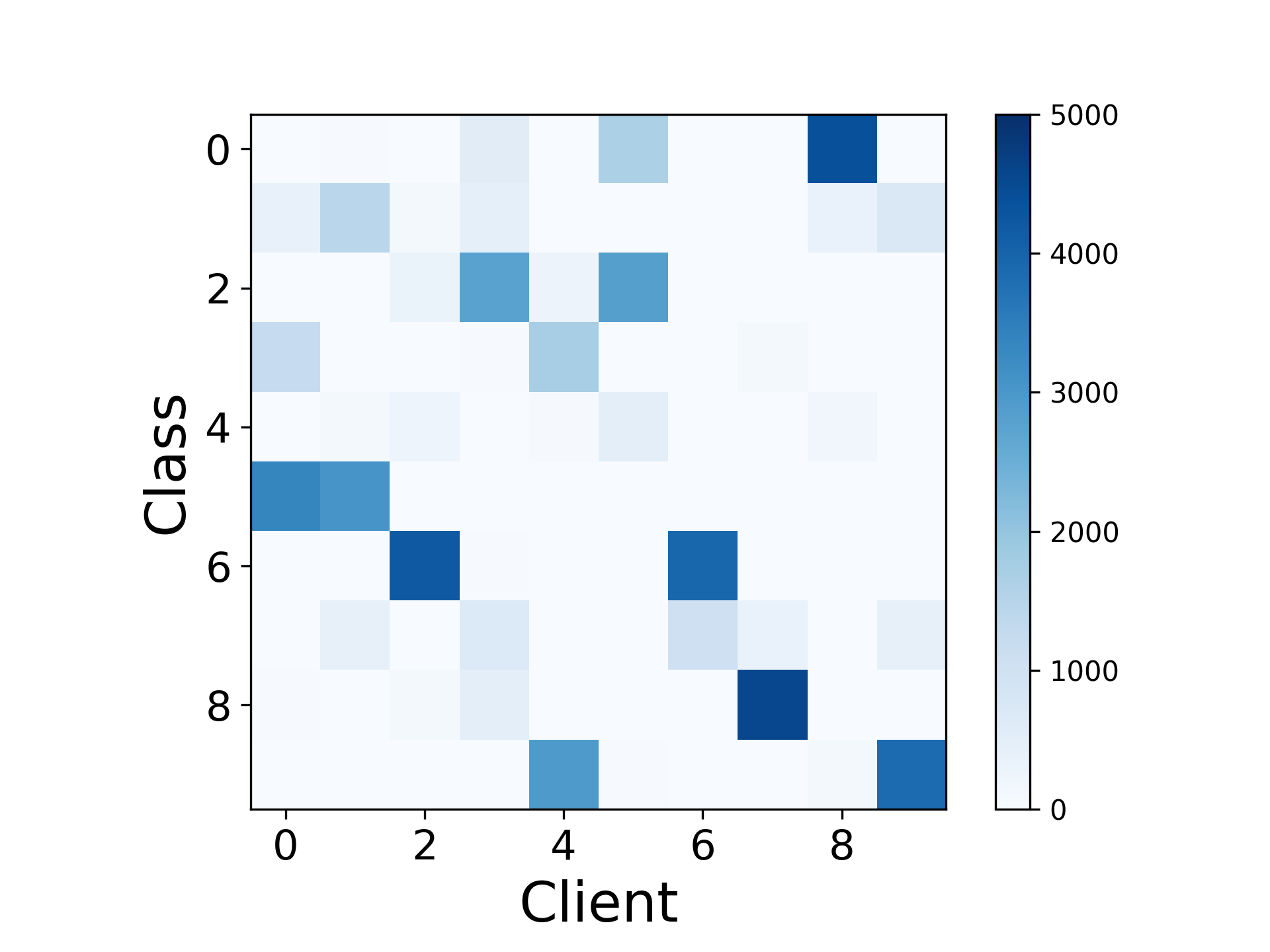}
         \caption{$\alpha=0.1$}
         \label{fig:y equals x}
     \end{subfigure}
     \begin{subfigure}[]{0.49\columnwidth}
         \centering
         \includegraphics[width=\columnwidth]{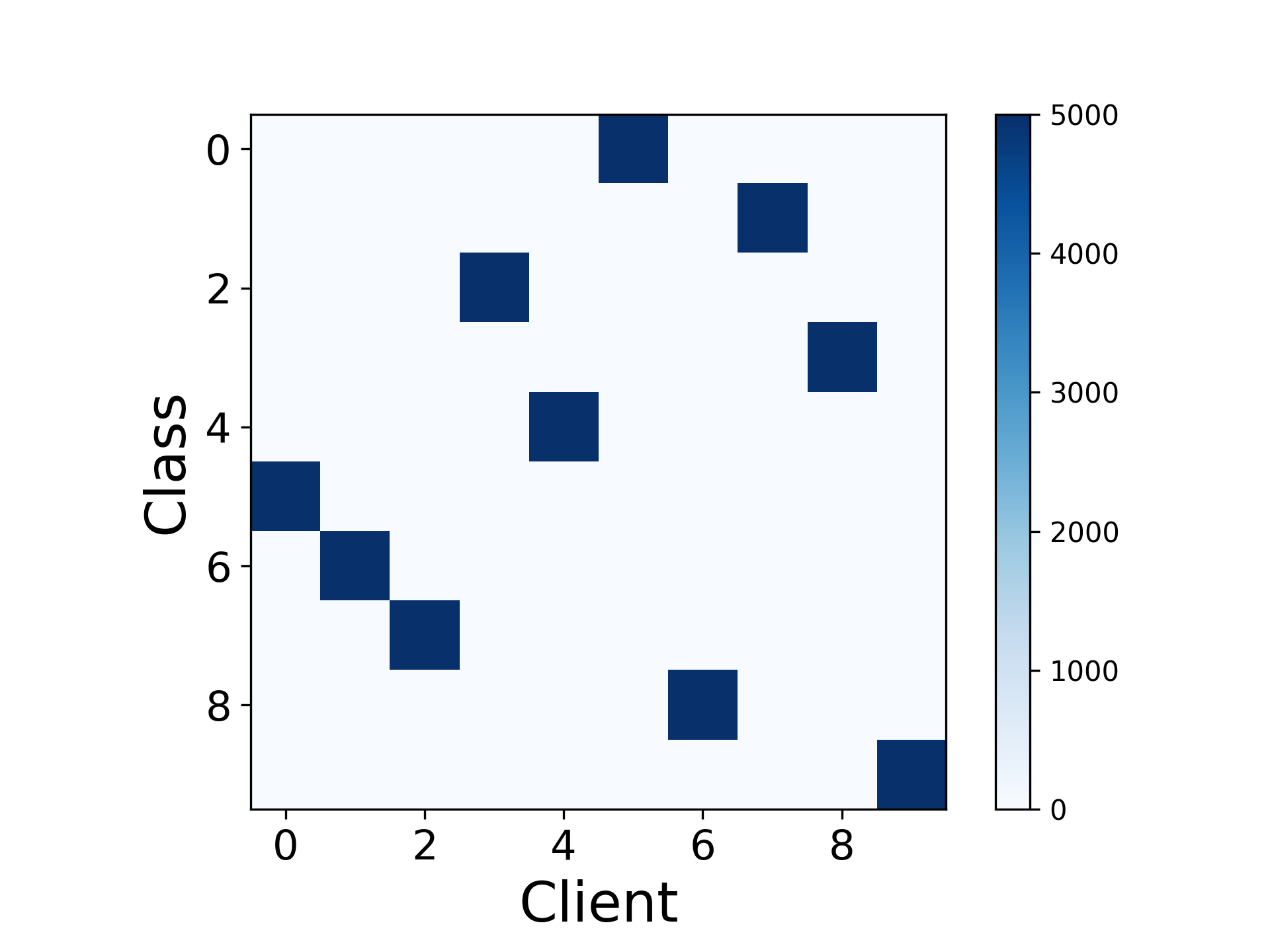}
         \caption{$\alpha=0.001$}
         \label{fig:three sin x}
     \end{subfigure}
        \caption{$Dir(\alpha)$ data partitioning for 10 clients on CIFAR-10. We show moderate ($\alpha=0.1$) to severe ($\alpha=0.01$) data heterogeneity levels. Data heterogeneity poses a significant challenge for many one-shot FL methods, as reconciling various models trained on widely different distributions is non-trivial. Our \method~approach rather trains diffusion models on the simple client distributions, which can then generate useful synthetic data for training global models.}
        \label{fig:hetero}
\end{figure}

The FashionMNIST dataset is an alternative to the original MNIST dataset, providing a more challenging task by replacing the handwritten digits with grayscale images of various fashion items. The dataset consists of 60,000 training images and 10,000 test images. 
The PathMNIST dataset is a medical dataset of colon pathology images in RGB, with a training set of 89,996 images and a test set containing 7,180 images with 9 classes.
The CIFAR-10 dataset consists of 60,000 color images equally distributed into ten different classes. 
The dataset is composed of a training set containing 50,000 images and a test set comprising of 10,000 images. CIFAR-10 is natively sized at 32$\times$32 pixels. We upsample FashionMNIST and PathMNIST from 28$\times$28 to 32$\times$32. A visualization of the dataset partitioning across clients is shown in Figure \ref{fig:hetero}.

We train with a batch size of 128 for all methods and use the AdamW optimizer. For local (and global training were applicable), we searched learning rates from [$3e^{-3}$, $1e^{-3}$, $3e^{-4}$, $1e^{-4}$] for each method using the CIFAR-10 dataset to find the optimal settings. We employed a ResNet16 architecture for the global model of all methods to ensure a fair comparison.
For DP experiments, we set the max gradient norm clipping threshold to 1.0 for all experiments and methods. In accordance with the recommendations of the Opacus \cite{opacus} library, we employ their Poisson batch sampling to ensure privacy guarantees.

As mentioned in Section 3.1 of the main paper, our DM is a basic U-Net structure with residual blocks \cite{ddpm, unet} and class-conditioning. Specifically, our U-Net has three downsampling stages (1/8 total downsampling) and three upsampling stages, each with residual convolutional blocks. A learnable embedding for time step and class conditioning is concatenated as additional input channels. For FedDiff$_{S}$, we halve the number of channels per block to reduce model size. For sampling at the server, we perform 1000 iterations as in \cite{ddpm} to generate each batch. The total number of generated samples is set equal to the size of the original dataset. Code available at \url{https://github.com/mmendiet/FedDiff}.

\begin{figure*}
     \centering
     \begin{subfigure}[]{0.32\textwidth}
         \centering
         \includegraphics[trim={0 30 0 0},clip,width=\textwidth]{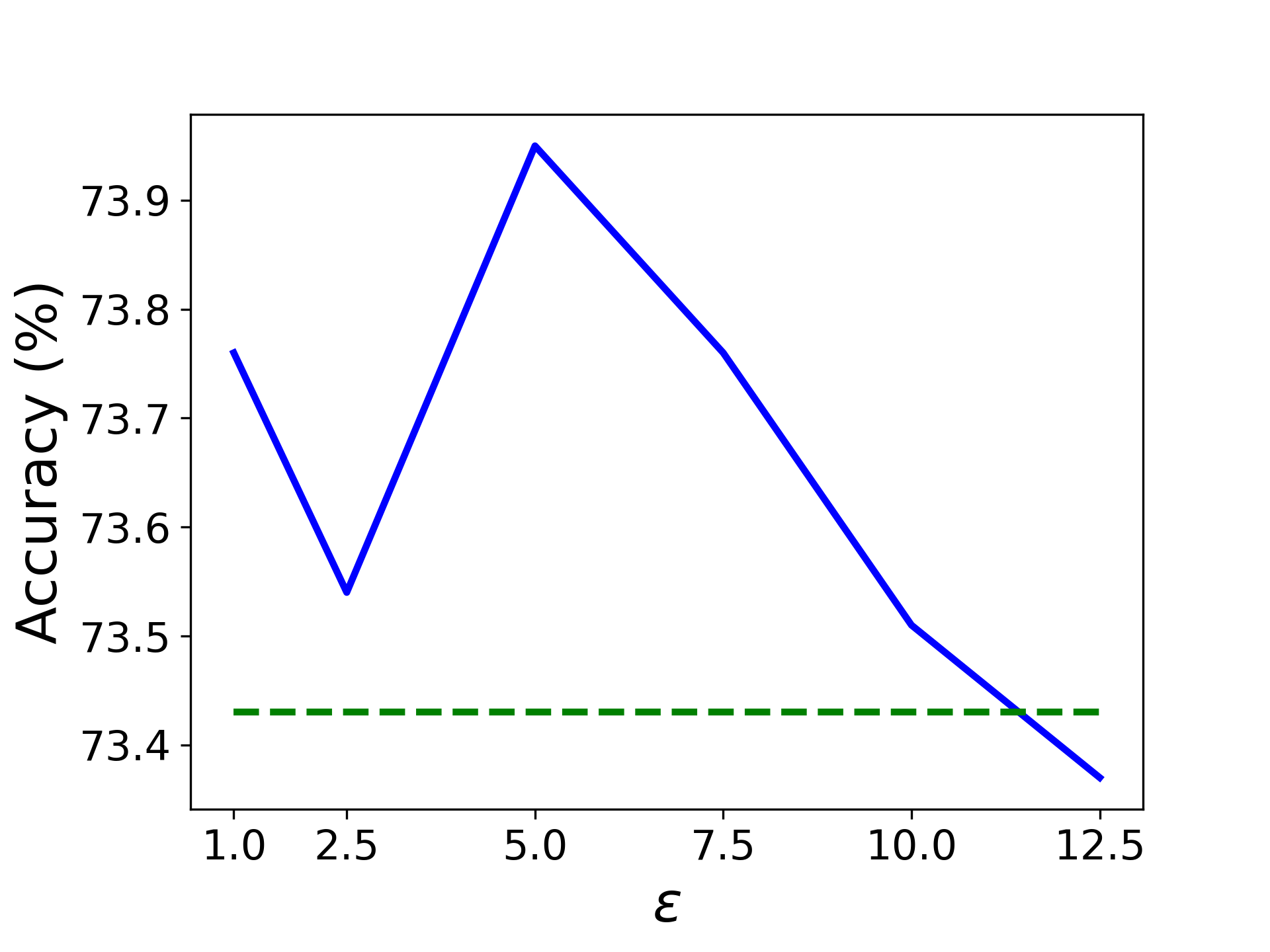}
         \caption{FashionMNIST}
     \end{subfigure}
     \hfill
      \begin{subfigure}[]{0.31\textwidth}
         \centering
         \includegraphics[trim={0 30 0 0},clip,width=\textwidth]{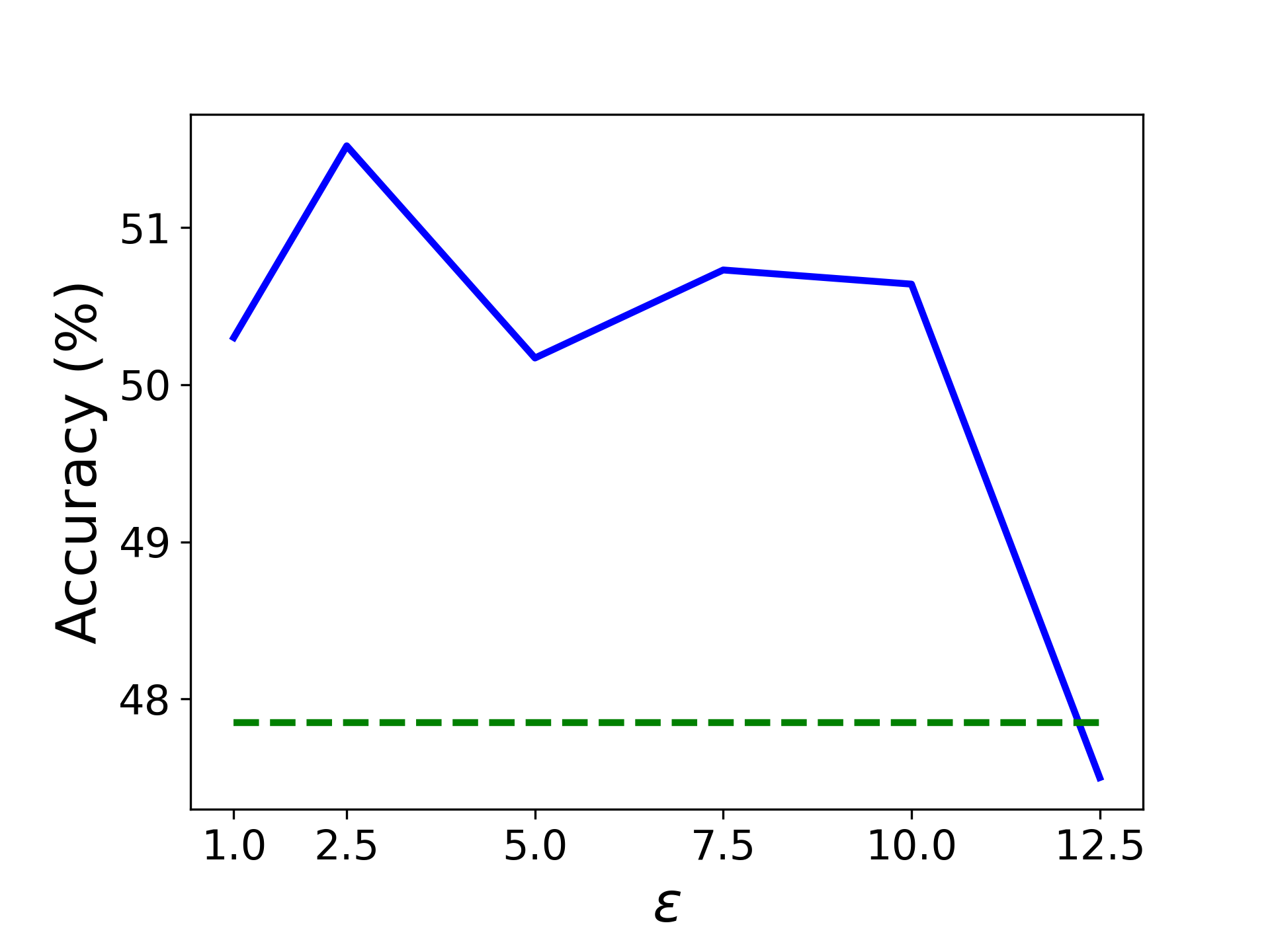}
         \caption{PathMNSIT}
     \end{subfigure}
     \hfill
     \begin{subfigure}[]{0.31\textwidth}
         \centering
         \includegraphics[trim={0 30 0 0},clip,width=\textwidth]{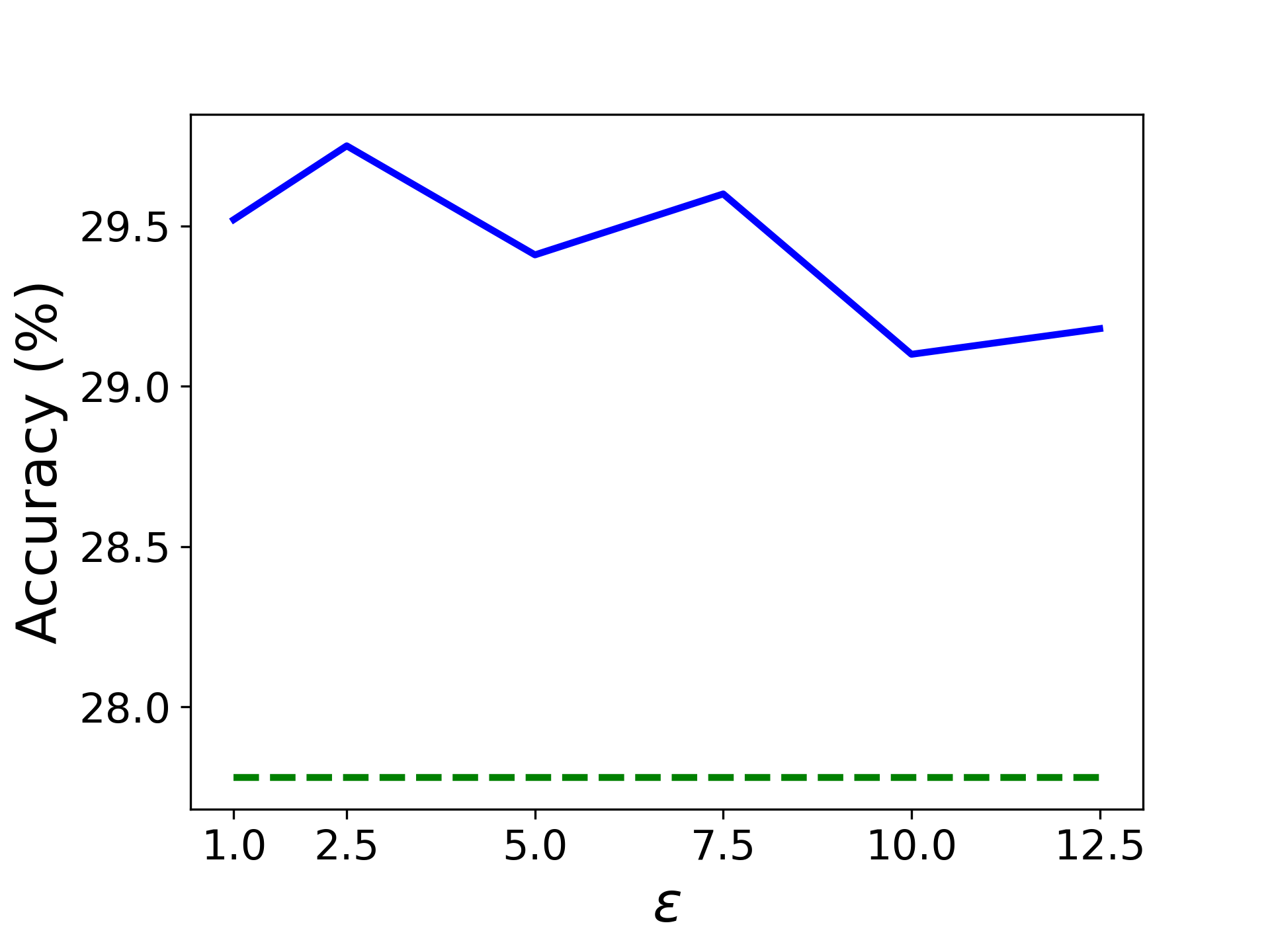}
         \caption{CIFAR-10}
     \end{subfigure}
        \caption{Ablation study of $\gamma$ in FMF under the $\epsilon$ = 10 setting. The accuracy of \method~is in \textbf{\textcolor{ForestGreen}{green}} and \method+FMF for various $\gamma$ in \textbf{\textcolor{blue}{blue}}. Generally, data filtering within the range of 1\% to 10\% produces positive outcomes, resulting in improved performance, with approximately 5\% serving as an effective default choice. We plot the mean across three runs with different seeds for each setting.}
        \label{fig:beta}
\end{figure*}

\section{Communication and Server-side Operations}
All methods primarily involve transmitting the model weights to the server, and this is done a single time. Additionally, specific information is sent. FedAvg and DENSE transmit the number of samples and label space, while FedCVAE and FedDiff send the number of samples per label. Thus, communication cost is mainly determined by the model size and number of clients. To ensure fairness between generative and discriminative methods, we select models with similar parameters and FLOPs, as shown in Table 3, maintaining comparable communication and computation costs.
We describe details of our DM architectures in Section \ref{sec:train}. As we also show in our experiments with FedDiff$_{S}$ in Table 3 of the main paper, we can adjust the size of the model to meet communication or compute needs and still provide exceptional performance.

Once the models are on the server, different operations are required for each method. DENSE, FedCVAE, and FedDiff require server-side generation and training, producing a dataset the same size as the original (e.g., 50k for CIFAR-10) and training a global model for 200 epochs. Server-side operation times on a single A5000 GPU for CIFAR-10 with 10 clients are 3.36, 1.76, and 2.17 hours for DENSE, FedCVAE, and FedDiff, respectively. DENSE takes the longest due to nested training of a GAN and the global model. Nonetheless, \ul{server computation is not a primary concern in FL}, as it is not constrained by the same resource limitations as client devices.

\section{Sampling Steps for Generation Ablation}\label{sec:sampling}
Diffusion models generate images through consecutive denoising steps, making the number of iterations per image a design choice when generating synthetic datasets at the server. Table \ref{tab:samp} provides insights into the impact of adjusting this parameter. Generally, increasing the number of steps enhances data quality and improves the final global model performance, particularly for the more difficult datasets. However, if generation time on the server is a concern, this parameter can be reduced, or increased when prioritizing data quality and model accuracy.
\begin{table}[h]
\small
\captionof{table}{Ablation on number of diffusion steps ($S$) used per image when generating the global synthetic dataset. Final global model accuracy is reported using our FedDiff approach under the default setting of $\alpha=0.01$ and $C=10$. The 1000-step setting is employed across all other experiments in the paper, as it is the standard practice in DDPM \cite{ddpm}.}
\label{tab:samp}
\centering
\resizebox{0.9\columnwidth}{!}{%
\begin{tabular}{ccccc}
\toprule
 Dataset & $S=100$ & $S=500$ & $S=1000$ & $S=2000$\\
\toprule
\multirow{1}{*}{FashionMNIST}
  & 86.49$\pm$0.23& 86.63$\pm$0.38&86.81$\pm$0.54 &86.75$\pm$0.17\\
\midrule
\multirow{1}{*}{PathMNIST}
  &69.85$\pm$1.15 &70.13$\pm$1.61 & 70.61$\pm$1.37 &71.44$\pm$1.83\\
\midrule
\multirow{1}{*}{CIFAR-10}
  &55.89$\pm$1.93 & 56.13$\pm$1.74 & 56.57$\pm$2.42 & 57.74$\pm$2.86\\
\bottomrule

\end{tabular}%
}
\end{table}

\section{FMF $\gamma$ Ablation} \label{sec:beta}
In Figure \ref{fig:beta}, we present the outcomes obtained using \method+FMF under $\epsilon=10$ across a range of $\gamma$ values, encompassing data filtering percentages spanning from 1\% to 12\%. Our findings indicate that, in general, data filtering within the 1\% to 10\% range yields favorable results and leads to performance enhancements, with around 5\% being a great default. Interestingly, the degree of improvement provided by FMF becomes more pronounced and consistent as the dataset becomes more challenging. This phenomenon aligns with the anticipated trends, as more intricate datasets inherently pose a greater challenge, making it less likely for the generators to consistently produce high-quality samples. Consequently, the need for data filtering becomes more pronounced in such scenarios to enhance sample quality.
This trend is also favorable since it addresses the specific need for improvement, especially in cases where performance is suboptimal and the challenges are more pronounced.

\section{Discussions, Limitations and Broader Impact} \label{sec:lim}

\textbf{Model Heterogeneity}. In real FL systems, model heterogeneity may often occur \cite{dense, fedcvae}. For instance, some clients may have architecture variations in their models or have smaller or larger models depending on their computing capabilities. Therefore, clients may have different architectures of similar generation capability, or even differing capabilities depending on the requirements of each client. Our approach allows for flexibility to accommodate such system diversity across clients. In \method, we generate data from the client models and employ that synthetic data for global training, and therefore can leverage varying models without the worry of reconciling the weights themselves. 

\textbf{Limitations and Broader Impact}.
One downside of our method is that the generated data, particularly under DP constraints, still lacks in quality and effectiveness for global model training versus using true data. For instance, with DP on CIFAR-10 as shown in Figure 4 in the main paper, the data loses a substantial amount of structure. An interesting direction for future work would be to study how to further improve the quality of the generated data and its usefulness for global model training while maintaining privacy. 
For instance, as differential privacy algorithms improve, FedDiff and the generated data quality will likewise benefit. Additionally, potentially leveraging prior information could allow the models to focus on task-relevant features, excluding irrelevant ones. By training a separate model to identify important features and then learning to generate images based only on these features with a DP secure model, we could potentially simplify the information needed for the generation process. This would likely lead to quicker convergence, allowing for better learning in DP settings with the same privacy budget and ultimately improving the privacy-utility trade-off of the generated samples.

Looking at the broader impact of our work, FL depends on the diversity of data contributed by different participants. If biases exist in the local datasets, they can be propagated and amplified during the model training process. This could lead to unintended algorithmic biases and discrimination in the resulting models. Ensuring diversity and fairness in the data used for FL is an important research direction to mitigate this risk and promote equitable outcomes \cite{bias}, particularly in the highly data heterogeneous environments explored in this work. 
Furthermore, as we have discussed throughout our paper, the privacy of client data is important in FL. To mitigate risks in this regard, we take many precautions to preserve privacy of the clients participated in the FL process though the use of DP, and operating within the one-shot setting to reduce the chance of eavesdropping.

\end{document}